\documentclass[11pt,a4paper]{article}
\usepackage[hyperref]{emnlp-ijcnlp-2019}
\usepackage{times}
\usepackage{latexsym}

\usepackage{url}

\usepackage{soul}
\usepackage{boxedminipage}
\usepackage{xcolor}
\usepackage{amsmath}
\usepackage{bm}
\usepackage{xspace}
\usepackage{enumitem}
\usepackage{cuted}

\aclfinalcopy % Uncomment this line for the final submission

\usepackage{multirow}
\usepackage{booktabs}

\usepackage{graphicx} %package to manage images
\graphicspath{ {figures/} }
\usepackage{subcaption}
\usepackage{rotating}
\usepackage[normalem]{ulem}

\usepackage{txfonts}

\DeclareMathOperator{\dec}{dec}
\DeclareMathOperator{\enca}{enc^x}
\DeclareMathOperator{\encb}{enc^t}

\DeclareMathOperator*{\softmax}{softmax}
\DeclareMathOperator*{\mlp}{MLP}

\newcommand{\ascal}{\alpha}

\newcommand{\xvec}{x}
\newcommand{\avec}{\mathbf{\alpha}}
\newcommand{\tvec}{t}
\newcommand{\hvec}{\mathbf{h}}
\newcommand{\vvec}{\mathbf{v}}
\newcommand{\cvec}{\mathbf{c}}
\newcommand{\svec}{\mathbf{s}}
\newcommand{\yvec}{y}
\newcommand{\step}[1]{_{#1}}

\newcommand{\xst}{^x}
\newcommand{\tst}{^t}

\newcommand{\gnc}[2]{\textcolor{magenta}{\sout{#1} #2}}

\usepackage{tikz}
\usetikzlibrary{shapes.geometric}
\usetikzlibrary{decorations.pathreplacing}
\usetikzlibrary{patterns}
\usepackage{pgfplots}

\usepackage[T2A,LGR,T1]{fontenc}
\usepackage[utf8]{inputenc}
\usepackage[russian,greek,english]{babel}

\title{Pushing the Limits of Low-Resource Morphological Inflection}

\author{Antonios Anastasopoulos and Graham Neubig\\
Language Technologies Institute, Carnegie Mellon University\\
\texttt{\{aanastas,gneubig\}@cs.cmu.edu}}

\begin{document}
\maketitle

%\textlatin

\begin{abstract}
Recent years have seen exceptional strides in the task of automatic morphological inflection generation.
However, for a long tail of languages the necessary resources are hard to come by, and state-of-the-art neural methods that work well under higher resource settings perform poorly in the face of a paucity of data.
In response, we propose a battery of improvements that greatly improve performance under such low-resource conditions. First, we present a novel two-step attention architecture for the inflection decoder. 
In addition, we investigate the effects of cross-lingual transfer from single and multiple languages, as well as monolingual data hallucination.
The macro-averaged accuracy of our models outperforms the state-of-the-art by~$15$ percentage points.\footnote{Our code is available at \url{https://github.com/antonisa/inflection}.}
Also, we identify the crucial factors for success with cross-lingual transfer for morphological inflection: typological similarity and a common representation across languages.
\end{abstract}

\begin{figure}
\tikzset{seq/.style={draw=none,fill=gray!20}}
\tikzset{layer/.style={->,thick}}
\tikzset{label/.style={anchor=west,font={\footnotesize}}}
\tikzset{seqlabel/.style={font={\small}}}
\newcommand{\encoder}[3]{
\draw[seq] (-1.25,-0.25) rectangle (1.25,0.25);
\node[seqlabel] at (0,0) 
%{$\xmat$};
{$#3\step{1} \cdots #3\step{#2}$};
\draw[layer] (0,0.3) -- (0,0.7);
\node[seqlabel] at (0,0.5) [label] {encoder};
\draw[seq] (-1.25,0.75) rectangle (1.25,1.25);
\node[seqlabel] at (0,1) %{$\hmat$}; 
{$\hvec\step{1}#1 \cdots \hvec\step{#2}#1$};
}
\newcommand{\decoder}[1]{
\draw[seq] (-1,1.75) rectangle (1,2.25);
\node[seqlabel] at (0,2) %{$\cmat#1$}; 
{$\cvec\step{1}#1 \cdots \cvec\step{K#1}#1$};
\draw[layer] (0,2.3) -- (0,2.7);
\node[seqlabel] at (0,2.5) [label] {decoder};
\draw[seq] (-1,2.75) rectangle (1,3.25);
\node[seqlabel] at (0,3) %{$\smat#1$}; 
{$\svec\step{1}#1 \cdots \svec\step{K#1}#1$};
\draw[layer] (0,3.3) -- (0,3.7);
\node[seqlabel] at (0,3.5) [label] {softmax};
\draw[seq] (-1,3.75) rectangle (1,4.25);
\node[seqlabel] at (0,4) %{$P(\ymat#1)$}; 
{$P(\yvec\step{1}#1 \cdots \yvec\step{K#1}#1)$};
}

\begin{center}
\resizebox{0.9\hsize}{!}{
\begin{tabular}{cc}
& \texttt{a g u \`a} \\
\multicolumn{2}{c}{
\begin{tikzpicture}
\begin{scope}[xshift=-1.4cm]
\encoder{^t}{M}{\tvec}
\draw[seq] (-1.,1.75) rectangle (1,2.25);
\node[seqlabel] at (0,2) %{$\cmat#1$}; 
{$\cvec\step{1}^t \cdots \cvec\step{K}^t$};
\draw[layer] (0,1.3) -- (0,1.7);
\node at (0,1.55) [label,anchor=east] {attention};
\draw[layer] (0,2.3) -- (0,2.7);
\draw[seq] (-1,2.75) rectangle (1,3.25);
\node[seqlabel] at (0,3) {$\svec\step{1} \cdots \svec\step{K}$};
\end{scope}
\begin{scope}[xshift=1.4cm]
\encoder{^x}{N}{\xvec}
\draw[seq] (-1.,1.75) rectangle (1,2.25);
\node[seqlabel] at (0,2) %{$\cmat#1$}; 
{$\cvec\step{1}^x \cdots \cvec\step{K}^x$};
\draw[layer] (0,1.3) -- (0,1.7);
\node at (0,1.55) [label] {attention};
\draw[layer] (0,2.3) -- (0,2.7);
\draw[seq] (-1,2.75) rectangle (1,3.25);
\node[seqlabel] at (0,3) {$\svec'\step{1} \cdots \svec'\step{K}$};
\draw[layer] (0,3.3) -- (0,3.7);
\node[seqlabel] at (0,3.5) [label] {softmax};
\draw[seq] (-1,3.75) rectangle (1,4.25);
\node[seqlabel] at (0,4) %{$P(\ymat#1)$}; 
{$P(\yvec\step{1} \cdots \yvec\step{K})$};
\end{scope}
%\draw[seq] (-1,2.75) rectangle (1,3.25);
%\node[seqlabel] at (0,3) {$\svec\step{1} \cdots \svec\step{K}$};
\node[seqlabel] at (0.6,2.5) [label,anchor=east] {decoder};
\draw[layer] (-.4,3) -- (0.4,3);
%\node[seqlabel] at (0,4.5) [label] {softmax};
%\draw[seq] (-1,4.75) rectangle (1,5.25);
%\node[seqlabel] at (0,5) %{$P(\ymat#1)$}; 
%{$P(\yvec\step{1} \cdots \yvec\step{K})$};
\end{tikzpicture}
} \\
{\small \texttt{\ V PRS 2 PL IND}} & \texttt{a g u a r}\\[-.8em]
\end{tabular}%
}
\end{center}
\caption{Visualization of our proposed two-step attention architecture.
%\footnote{We only used a single layer for both the encoder and decoder. It is possible to use multiple stacked RNNs or self-attention layers; typically, the output of the encoder(s) and decoder(s) ($\cvec\step{n}$ and $P(\yvec\step{k})$, respectively) would be computed from the top layer only.}
The decoder first attends over the tag sequence $\mathbf{T}$ and then uses the updated decoder state $\svec'$ to attend over the character sequence $\mathbf{X}$ in order to produce the inflected form $\mathbf{Y}$. (Example from Asturian.)}
\label{fig:model}
\end{figure}
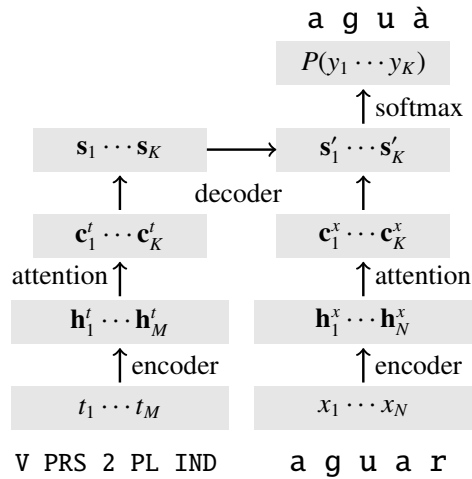

\section{Introduction}

The majority of the world's languages are categorized as synthetic, meaning that they have rich morphology, be it fusional, agglutinative, polysynthetic, or a mixture thereof.
As Natural Language Processing (NLP) keeps expanding its frontiers to encompass more and more languages, modeling of the grammatical functions that guide language generation is of utmost importance.

In the case of morphologically-rich languages, explicit modeling of the inflection processes has significant potential to alleviate issues created by data scarcity and the resulting lack of vocabulary coverage.
Especially on low-resource, under-represented languages and dialects, the potential for \textit{impact} is much higher.
For example, speech recognition \cite{foley2018building} and predictive keyboards \cite{breiner2019automatic} for under-represented languages, if they exist, largely still rely on unigram lexicons with performance inferior to the sophisticated language models used in high-resource ones.
Good inflection models would be invaluable for predictive text technology in morphologically-rich languages as they could effectively enable proper handling of the huge vocabulary.

Additionally, they could be very useful for building educational applications for languages of under-represented communities (along with their inverse, morphological analyzers). Encouraging examples are the Yupik morphological analyzer \cite{schwartz2019bootstrapping} and the Inuktitut educational tools from the respective Native Peoples communities.\footnote{\url{http://www.inuktitutcomputing.ca/}}
%in Alaska and Canada.
The social impact of such applications can be enormous, effectively raising the status of the languages slightly closer to the level of the dominant regional language. 

Morphological inflection has been thoroughly studied in monolingual high resource settings, especially through the recent SIGMORPHON challenges \cite{cotterell2016sigmorphon,cotterell-etal-2017-conll,cotterell-etal-2018-conll}. Low-resource settings, in contrast, are relatively under-explored. One promising direction and the main focus of the SIGMORPHON 2019 challenge \cite{mccarthy-etal-2019-sigmorphon} is cross-lingual training, which has been successfully applied in other low-resource tasks such as Machine Translation (MT) or parsing.

In this work we focus in this cross-lingual setting for low-resource morphological inflection and propose several simple, yet effective approaches to mitigating problems caused by extreme lack of data which, put together, \emph{improve accuracy by 15 percentage points over a state-of-the-art baseline}. 
This is achieved through the combination of a novel decoder architecture, a training regime that alleviates the need for costly structural biases that force attention monotonicity, and a data hallucination technique. We also present thorough ablations and identify the crucial factors for success with our approach.

Our system was the best performing system in the 2019 SIGMORPHON shared task on morphological inflection when evaluated on accuracy, while it ranked third when evaluated with average Levenshtein distance.

\section{Task Definition and Approach}
%\an{2CHECK}
Morphological inflection is the process that creates grammatical forms (typically guided by sentence structure) of a lexeme/lemma.
As a computational task it is framed as mapping from the lemma and a set of morphological tags to the desired form, which simplifies the task by removing the necessity to infer the form from context. For an example from Asturian, given the lemma \texttt{aguar} and tags {\small\texttt{V;PRS;2;PL;IND}}, the task is to create the indicative voice, present tense, 2\textsuperscript{nd} person plural form~\texttt{agu\`{a}}.

Let $\mathbf{X}=\xvec\step{1}\ldots\xvec\step{N}$ be a character sequence of the lemma, $\mathbf{T}=\tvec\step{1}\ldots\tvec\step{M}$ a set of morphological tags, and $\mathbf{Y}=\yvec\step{1}\ldots\yvec\step{K}$ be an inflection target character sequence. 
The goal is to model~$P(\mathbf{Y}\mid\mathbf{X},\mathbf{T})$.

Our approach consists of three major components. 
First, we propose a novel two-step attention decoder architecture (\S\ref{sec:model}). Second, we augment the low-resource datasets with a data hallucination technique (\S\ref{sec:hallucination}).
Third, we devise a training schedule (\S\ref{sec:training}) that substitutes structural biases for attention monotonicity.
%Last, in Section~\ref{sec:inference} we outline how inference is performed and list our model's hyperparameter details.

\subsection{Model Architecture}
\label{sec:model}

Our models are based on a sequence-to-sequence model with attention~\cite{bahdanau2015}. In broad terms, the model is composed of three parts: an encoder, the attention, and a recurrent decoder. In the setting of the inflection task, there is an additional input provided (the set of morphological tags) which requires an additional encoder.

A visualization of our model is shown in Figure~\ref{fig:model}.
First, an encoder transforms the input character sequence $\xvec\step{1}\ldots\xvec\step{N}$ into a sequence of intermediate representations $\hvec^x\step{1}\ldots\hvec^x\step{N}$. A second encoder transforms the set of morphological tags $\tvec\step{1}\ldots\tvec\step{M}$ into another sequence of states~$\hvec\tst\step{1}\ldots\hvec\tst\step{M}$:
{\setlength{\abovedisplayskip}{4pt}
\setlength{\belowdisplayskip}{4pt}
\begin{align*}
    \hvec\xst\step{n} &= \enca(\hvec\xst\step{n-1}, \xvec\step{n}) &\text{and} &&\hvec\tst\step{m} &= \encb(\mathbf{T}).
\end{align*}}\noindent
In our implementation, we use a single layer bi-directional recurrent encoder for the lemma, and a self-attention encoder \cite{transformer} for encoding the tags as there is no inherent order (e.g. left-to-right) in their presentation. In preliminary experiments, a self-attention lemma encoder proved hard to train, while a recurrent tag encoder yielded quite competitive results.

Next, we have attention mechanisms that transform the two sequences of input states into two sequences of \emph{context vectors} via two matrices of \emph{attention weights} ($k$ is the current decoder time step):
{\setlength{\abovedisplayskip}{4pt}
\setlength{\belowdisplayskip}{4pt}
\begin{align*}
    \cvec\xst\step{k} &= \begin{bmatrix} \sum_n \ascal\xst_{kn} \hvec\xst\step{n} \end{bmatrix}
    &\cvec\tst\step{k} &= \begin{bmatrix} \sum_m \ascal\tst_{km} \hvec\tst\step{m} \end{bmatrix}.
\end{align*}}\noindent
Finally, the recurrent decoder computes a sequence of \emph{output states} in a two-step process, from which a probability distribution over output characters can be computed:
{\setlength{\abovedisplayskip}{4pt}
\setlength{\belowdisplayskip}{4pt}
\begin{align*}
\svec\step{k} &= \svec'\step{k-1} + \cvec\tst\step{k} \\
\svec'\step{k} &= \dec(\svec'\step{k-1}, \cvec\xst\step{k}, \yvec\step{k-1}) \\
P(\yvec\step{k}) &= \softmax(\svec'\step{k}).
\end{align*}}\noindent
The attention mechanisms produce their weights as in~\citet{luong2015effective}, with $\vvec\xst$, $\vvec\tst$, $\mathbf{W}_{\avec\xst}^{s'}$, $\mathbf{W}_{\avec\tst}^s$, $\mathbf{W}_{\avec\xst}^h$, and $\mathbf{W}^h_{\avec\tst}$ being parameters to be learned:
{\setlength{\abovedisplayskip}{4pt}
\setlength{\belowdisplayskip}{4pt}
\begin{align*}
\mathbf{\ascal}\tst\step{km} &= \softmax (\vvec\tst \tanh (\left[ \mathbf{W}_{\avec\tst}^{s} \svec'\step{k-1} ; \mathbf{W}_{\avec\tst}^h \hvec\tst\step{m} \right]) )\\
\mathbf{\ascal}\xst\step{kn} &= \softmax (\vvec\xst \tanh (\left[ \mathbf{W}_{\avec\xst}^{s'} \svec\step{k} ; \mathbf{W}_{\avec\xst}^h \hvec\xst\step{n} \right]) ).
\end{align*}}\noindent
The two-step attention process essentially first uses the decoder's previous state $\svec'\step{k-1}$ as the query for attending over the tags. Then, it creates a tag-informed state $\svec'_k$ by adding the tag context $\cvec^t\step{k}$ to the previous state. The tag-informed state is then used as the query for attending over the source characters and produce the context $\cvec^x\step{k}$. The last step is then to update the recurrent state and produce the output character.
Ultimately, we desire that the provided tag set guides the generation, which also means influencing the attention over the characters of the lemma.

\paragraph{Additional Structural Biases for Attention} Incorporating structural biases in the model's architecture or in the training objective can lead to improvements in performance, especially for tasks where the attention mechanism is expected to behave similarly to an alignment model, like MT. This idea has been successfully applied to the inflection task
%\cite{aharoni2017morphological}
and is at the core of the state-of-the-art model~\cite{wu-2019-exact}.

One bias we deem important is \emph{coverage} of all input characters and tags from the attentions.
Intuitively, this entails encouraging the model to ``look at'' the whole input. We take the approach of \citet{cohn2016incorporating} and add two regularizers over the final attention matrices, encouraging them to
also
sum to one column-wise: 
{\setlength{\abovedisplayskip}{4pt}
\setlength{\belowdisplayskip}{4pt}
\begin{align*}
	-&\lambda \parallel \Sigma_j a^t_{jm}  - \mathbb{I} \parallel_2
	&-\lambda \parallel \Sigma_j a^x_{jn}  - \mathbb{I} \parallel_2
\end{align*}
}\noindent
Another bias that we incorporate encourages the Markov assumption over attention/alignments. Briefly, this means that if the $i$-th source character/word is aligned to the $j$-th target one ($i \leftarrow j$), then alignments $i$+$1$$\leftarrow$$j$+$1$ or $i$$\leftarrow$$j$+$1$ are also quite likely. In a neural architecture this can be approximated by providing the attention weight vector from the previous timestep as
%additional 
input to the function that computes the attention weights. We refer the reader to \citet{cohn2016incorporating} for exact details.

\paragraph{Adversarial Language Discriminator} When training multilingual systems, encouraging the encoder to learn language-invariant representations can often lead to improvements \cite{xie2017controllable,chen2018adversarial},
as it forces the model to truly work in a multilingual setting.
We achieve that by introducing a language discriminator \cite{ganin2016domain}. This additional component receives the last output of the (bi-directional) intermediate lemma representations $\hvec^x_N$ and outputs a prediction $y_l$ of the source language such that $y_l$=$\softmax(\mlp(\hvec^x_N)).$

The discriminator is trained to predict the language by minimizing a standard cross-entropy loss~$\mathcal{L}_l$ similar to \citet{lample2017unsupervised}. However, in order to encourage the encoder to learn language-invariant representations,  we reverse the gradients flowing from that component into the encoder during back-propagation.

\begin{figure}[t]
    \centering
    \begin{tabular}{lc}
        \multicolumn{1}{l}{Original triple} & 
         \begin{tikzpicture}
         \draw[decorate,decoration={brace,amplitude=3pt}] (-0.2,0) -- (0.7,0) node[midway,yshift=1em]{\textit{stem}};
        \draw[decorate,decoration={brace,amplitude=3pt}] (1.7,0) -- (2.5,0) node[midway,yshift=1em]{\textit{stem}};
        \end{tikzpicture} \ \ \ \ \\
        \ \ \ lemma & { \foreignlanguage{greek}{π α ρ α κ ά μ π τ ω \ }}\\
         & 
         \begin{tikzpicture}
            \draw[thick,red] (0,0) -- (0,0.7);
            \draw[thick,red] (0.33,0) -- (0.33,0.7);
            \draw[thick,red] (0.65,0) -- (0.65,0.7);
            \draw[thick,dashed] (0.95,0) -- (0.95,0.7);
            \draw[thick] (1.25,0) -- (1.25,0.7);
            \draw[thick,dashed] (1.55,0) -- (1.55,0.7);
            \draw[thick,red] (1.88,0) -- (1.88,0.7);
            \draw[thick,red] (2.2,0) -- (2.2,0.7);
            \draw[thick,red] (2.45,0) -- (2.5,0.7);
            \draw[thick,dashed] (2.7,0) -- (2.78,0.7);
            \draw[thick,dashed] (2.9,0) -- (2.78,0.7);
         \end{tikzpicture}
         \\
        \ \ \ {\small\texttt{+V;2;SG;IPFV;PST}} & {\foreignlanguage{greek}{π α ρ έ κ α μ π τ ε ς}}\\
        \midrule
        \multicolumn{1}{l}{Hallucinated} & \\
        \ \ \ lemma & { \foreignlanguage{greek}{\textcolor{red}{π ξ ρ} α κ ά \textcolor{red}{μ ο τ} ω \ \ }}\\
        \ \ \ {\small\texttt{+V;2;SG;IPFV;PST}} & {\foreignlanguage{greek}{\textcolor{red}{π ξ ρ} έ κ α \textcolor{red}{μ ο τ} ε ς}} \\
    \end{tabular}
    \caption{Example of our hallucination process (Greek). The lemma and inflected forms are aligned at the character level. The inside of \textit{stem}-considered parts (highlighted) are substituted with random characters, creating hallucinated triples (bottom). }
    \label{tab:hall_example}
\end{figure}

\subsection{Data Hallucination}
\label{sec:hallucination}

Low-resource language datasets are usually too small to allow for proper learning with neural networks. 
%As we show in our analysis in Section~\ref{sec:analysis}, training only with the provided datasets leads to several issues related to exposure and label bias \gn{``exposure bias and label bias'' may not be universally recognized terms, so it might be good to mention these in plainer terminology?}.
A major issue in our case is \textit{label bias}. Put simply, the character decoder will overly prefer outputting 
%the more
common character sequences. However, with just~50 examples to learn from, the output character n-gram distribution will hardly match the real one, because the majority of the n-grams will have zero probability.
In order to mitigate this issue, we augment our training sets with hallucinated data. 

%As several previous works have noted, 
In most languages morphological reinflection is realized by adding, deleting, or modifying prefixes or suffixes over a stem that is mainly unchanged.
Though we do not have prior information regarding the stem or affixes of our data we can use character alignment to approximately compute them.
We use the alignment method from the {SIGMORPHON} 2016 baseline \cite{cotterell2016sigmorphon} to align the lemmata and the inflected forms. 
For each example, we consider as part of the stem \textit{any} sequence of three or more\footnote{This heuristic number is largely arbitrary and could potentially be tuned to different values for each language.} consecutive lemma characters that are aligned to the exact same characters in the inflected form.

Now, for each such region considered as a ``stem'', we randomly substitute its inside (not start or end) characters with other characters from the language's alphabet. Note that we do not change the length of the region, though allowing for such variation could possibly lead to further improvement.
The substitution characters are sampled uniformly for the alphabet, rather than attempting to sample from a more informed distribution, which
% . Although \citet{silfverberg2017data} have noted otherwise, we believe that sampling from a more informed distribution
has potential for further improvements. 
%We further explain why this strategy is preferable for low-resource languages in Section~\ref{sec:analysis}.
Overall, we hallucinated 10,000 examples for each low-resource language, creating an additional hallucinated dataset $\mathcal{H}$.

A visualized example of our hallucination process is outlined in Figure~\ref{tab:hall_example}. Out of the three regions with matching aligned characters (thick lines), we identify two with length equal to three or more. In the hallucinated example (bottom of the figure), we sample random characters for the inside of such regions.

\citet{silfverberg2017data} have proposed a data hallucination method conceptually quite similar to ours, which treats the \textit{single} longest common continuous substring between lemma and form as the stem.
%and randomly samples characters to replace it.
Their approach would be effectively similar to ours for languages with affixal stem-invariant morphology,
%(e.g. agglutinative)
but it would likely fail in more complicated morphological phenomena like apophony, stem alternation 
(conceptually similar to infix morphology) 
or in the root-and-pattern morphology of semitic languages.
Consider the apophony example from the past participle form \texttt{gschwommen} of the lemma \texttt{schwimmen} in Swiss German: our approach would treat both the \texttt{schw} and the \texttt{mmen} regions as a stem, as opposed to only considering one.\footnote{Most likely, the desired parts are \texttt{schw} and \texttt{mm}.}
Note that neither of the two approaches are suitable for phenomena such as suppletion (e.g. the inflection of the Spanish verb \texttt{ir} `to go' into \texttt{fue} `went') or phenomena like the reduplication pattern of Indonesian noun plurals as in \texttt{kuda} `horse', \texttt{kuda-kuda} `horses' \cite{sneddon2012indonesian}.

\begin{table}[t]
    \centering
    \begin{tabular}{@{}l@{}cc@{}}
    \toprule
        \multicolumn{1}{c}{\multirow{1}{*}{Model}} &  Accuracy & Median \\
    \midrule
    \citet{wu-2019-exact} & 48.5 & 45.5\\
    %simple multi-source & 40.1 \\
    this work & 48.8 & 67.0\\
    \ \ \ \ $+ \mathcal{H}$ & 60.1 & 66.6 \\
    \ \ \ \ $+ \mathcal{H} + \mathcal{L}_l$ & 60.8 & 66.0 \\
    \ \ \ \ $+$multi-language transfer & \textbf{63.8} & 64.0\\
    \midrule
    oracle & 68.2 & 74.0\\
    \bottomrule
    \end{tabular}
    \caption{Macro-averaged accuracy over~100 language test pairs. Our best model outperforms the baseline by $15$ percentage points.}
    \label{tab:test_summary}
\end{table}

\subsection{Training Schedule}
\label{sec:training}

Our training schedule attempts to balance two desires: helping the model to (1) learn to copy,  and (2) learn cross-lingually with a particular focus on the low-resource language. To achieve this, we split training into three phases: warm-up, cross-lingual training, and low-resource fine-tuning.

\paragraph{Phase 1: Warm-Up} As several previous works have noted,
%(see Section~\S\ref{sec:related}) 
learning to copy is crucial. Unlike other proposed models, though, our model does not include any structural biases that encourage copying. Instead, we rely on an additional copying task in a warm up period.

We transform each training triple $[\mathbf{X}, \mathbf{T}, \mathbf{Y}]$ into two additional triples that encourage copying: $[\mathbf{X}, \texttt{COPY}, \mathbf{X}]$ and $[\mathbf{Y}, \mathbf{T}, \mathbf{Y}]$.
Using both the input and output sequences for the copying task allows the use of slightly more diverse data. Note that we only use the correct tags when copying the inflected form. For copying the lemma we use a specialized tag (\texttt{COPY}). Additional improvements could be achieved if one knew the exact tags that match the lemmata. But this could vary by language: an English verb's lemma is its infinitive and has a {\small\texttt{V;NFIN}} tag, while a verb lemma in Modern Greek is its {\small\texttt{V;1;SG;IPFV;PRS}} form.

In this stage we use a relatively large batch size~(10) in order to encourage more coarse updates.
In most cases, the model achieves extremely high copying accuracy after a couple of warm-up epochs, so we stop the warm up stage when copying accuracy exceeds $75\%$. At this point the attention mechanism over the source characters has learned to be monotonic. In contrast, the model of \citet{wu-2019-exact} requires a dynamic programming method that forces strict monotonicity, with an additional (non-parallelizable) computational cost $\mathcal{O}(\vert x\vert^2)$ throughout training.

\paragraph{Phase 2: Cross-lingual training} In the main training phase we use both high- and low-resource language data (including any hallucinated data). If not using hallucinated data, we up-sample the low-resource data in order to match the size of the high-resource ones. Furthermore, with probability $0.30$ we also sample copying tasks to intersperse throughout the training epoch. 
This ensures the source-character attention keeps being monotonic.

\paragraph{Phase 3: Fine-tuning} The last phase is inspired by fine-tuning, or continued training, as applied to cross-lingual MT e.g. by \citet{neubig18emnlp}
or to domain adaptation for MT e.g. by \citet{luongstanford}. 
The setting is nearly identical to the second phase, except we only use the low-resource language data for training and do not use the copying task. Furthermore, we substitute teacher forcing for scheduled sampling \cite{bengio2015scheduled}, where with probability $50\%$ the input to the next step of the decoder is not the gold one, but its previous prediction. This technique allows the model to become more robust to its own mistakes, effectively limiting the effect of exposure bias.
It is worth noting that at this point, the learning rate is typically quite small, so we reduce the batch size to a single instance. In most cases, though, the improvements on development set accuracy are marginal~($1$ to~$2\%$).

\begin{table}[t]
    \centering
    \begin{tabular}{@{}l|c@{}}
    \toprule
        \multicolumn{1}{c|}{\multirow{2}{*}{Model}} &  Dev Accuracy\\
        & \small{(macro-averaged)} \\
    \midrule
    \citet{wu-2019-exact}: & \\
    \ \ \ \ 0\textsuperscript{th}-order soft attention & 29.6 \\
    \ \ \ \ 0\textsuperscript{st}-order hard attention & 32.3 \\
    \ \ \ \ 0\textsuperscript{th}-order monotonic attention & 37.2 \\
    \ \ \ \ 1\textsuperscript{st}-order monotonic attention & 40.3 \\
    \midrule
    %simple multi-source & 40.1 \\
    two-step attention (this work) & \\
    \ \ \ \ $-$ warm-up, copy task & 31.2 \\
    \ \ \ \ $+$ warm-up, copy task & 48.0 \\
    \ \ \ \ $+$ structural biases & 48.7 \\ 
    \ \ \ \ $+$ scheduled sampling & 49.1 \\
    \ifaclfinal
    \ \ \ \ $+$ minibatch schedule & 49.4 \\
    \fi
    \midrule 
    with hallucinated data: \\
    \ \ \ \ $+ \mathcal{H}$ & $82.6^\dagger$ \\
    \ \ \ \ $+ \mathcal{H} + \mathcal{L}_l$ & $85.4^\dagger$ \\
    \ \ \ \ $+$ensemble (three models) & $85.6^\dagger$\\
    \bottomrule
    \end{tabular}
    \caption{Our proposed two-step attention architecture outperforms the baselines. Additional biases and scheduled sampling contribute further improvements. $\dagger$: results with hallucinated data are not directly comparable, as dev data were used for hallucination.}
    \label{tab:model_ablations}
\end{table}
\begin{table*}[t]
\centering
\small
\begin{tabular}{cccccccc}
\toprule
L1 & L2 & Genetic dist. & L1+L2 & $+\mathcal{L}_l$ & $+\mathcal{H}$ & $+\mathcal{L}_l+\mathcal{H}$ & $\mathcal{H}$ \\ \midrule
latin & czech & 0.86 &  15.0 &  26.0 &  71.4 &  68.0 &  \textbf{77.4} \\ 
bengali & greek & 0.88 &  22.4 &  16.4 &  70.5 &  70.6 &  \textbf{71.6} \\
sorani & irish & 1.0 &  20.3 & 18.6 &  \textbf{66.3} &  64.6 &  65.6 \\
italian & ladin & 0.30 &  48 &  54 &  \textbf{74} &  \textbf{74} &  \textbf{74}  \\
latvian & lithuanian & 0.25 &  17.1 &  23.2 &  48.4 &  48.4 &  \textbf{50.5}  \\
english & murrinhpatha & N/A &  \textbf{36} &  2 &  6 &  20 &  20  \\
italian & neapolitan & 0.10 & 70 &  70 &  83 &  83 &  \textbf{84} \\ 
urdu & old english & 0.88 & 23.8 &  18.5 &  43.4 &  40.4 &  \textbf{44.3} \\
slovene & old saxon & 0.89 & 10.7 &  14.7 &  \textbf{52.3} &  50.5 &  50.5  \\
russian & portuguese & 0.93 & 34.5 &  22.2 &  \textbf{88.8} &  88.4 &  87.7  \\
swahili & quechua & 1.0 & 14.2 &  12.8 &  \textbf{92.1} &  91.6 &  91.6  \\ 
portuguese & russian & 0.93 &  25.6 &  17.5 &  \textbf{76.3} &  74.6 &  74.3  \\
kurmanji & sorani & N/A & 16.2 &  13.6 &  \textbf{69.0} &  64.3 &  66.7 \\
zulu & swahili & 1.0 & 46 &  52 &  \textbf{81} &  80 &  76 \\
kannada & telugu & 0.75 & 76 &  72 &  \textbf{94} &  90 &  \textbf{94} \\ \midrule
\multicolumn{2}{c}{Average} & 0.75 & 31.72 & 28.90 & 67.77 & 67.23 & 68.55 \\
\bottomrule
\end{tabular}
\caption{Results with a single transfer language. Monolingual data hallucination is crucial due to the distance of the languages. In some cases cross lingual transfer should be avoided in favor of a purely monolingual setting ($\mathcal{H})$.}
\label{tab:single_results}
\end{table*}

\subsection{Inference and Implementation Details}
\label{sec:inference}

%We treat training with the language adversarial loss as an additional hyperparameter (decision) so we choose models trained with or without it based on their development set performance.

The performance of inflection systems is typically evaluated with exact-match token-level accuracy, as well as character-level Levenshtein distance.\footnote{Due to space constraints, we do not provide Levenshtein distance results. They generally inversely correlate with token-level accuracy.} Hence, during training we continuously evaluate the model's performance on the development set with both metrics. Consequently we store three checkpoints: 
%as our final models:
the one that achieved the highest accuracy, the one that reached the lowest Levenshtein distance, and the one that improved on both metrics over previous dev evaluations. In a few cases these three checkpoints coincide, but this is rather rare. We ensemble these three models with equal weights for producing our final predictions.

%\paragraph{Implementation} 
All our models are implemented in DyNet \cite{neubig2017dynet}. Each model is trained on a single CPU, as each training run requires less than~1GB of RAM and typically concludes within~3 to~4 hours.
We provide additional hyper-parameter details in the Appendix.

\section{Empirical Results}
\label{sec:results}

Our experiments are conducted on the SIGMORPHON 2019 challenge datasets \cite{sigmorphon2019}. The dataset consists of~100 pairs between (mostly related) high-resource transfer languages and 43 low-resource test languages with examples taken from the Unimorph database \cite{kirov-etal-2018-unimorph}. The low-resource languages typically have about~100 training examples, plus~50--100 (or in very few cases, 1000) examples in the development set.
In contrast, most high-resource language training sets include~10,000 instances.

\paragraph{Test Accuracy} Our main results are summarized in Table~\ref{tab:test_summary}. Our models significantly outperform the baseline, evaluated with macro-averaged accuracy over the~100 language pairs.
Our novel architecture performs slightly better than the baseline without any additional data. 
Importantly, including the hallucinated datasets boosts total accuracy by~10 percentage points, while transfer from multiple languages further improves to a state-of-the-art accuracy of~$63.8\%$ -- higher than any system submitted to the shared task.

It is worth noting that all of these improvements are not uniform across languages/pairs: without hallucinated training data, our model's median is notably larger than its average, implying that for a few language pairs our model is under-trained and under-performs (in particular, pairs with Yiddish, Votic, and Ingrian as test languages).
%We further discuss and expand on this observation in Section~\ref{sec:analysis}. 
Nonetheless, if one had access to an oracle that would allow them to select the best performing model out of all the different settings, they would achieve an oracle accuracy of~68.2\%. 

\paragraph{Architecture Ablations} We focus on the architecture of the model and the structural biases we introduced, using the development set for evaluation. Table~\ref{tab:model_ablations} presents the macro-averaged accuracy for the baseline \cite{wu-2019-exact} and the different versions of our model.

The first thing to note is the importance of the warm-up period in our training schedule and the use of the copying task. Our model neither handles copying in any explicit way nor encourages the attention to be monotonic. Without the warm-up period and the additional copying tasks, our model's development accuracy is in the same ballpark as the baseline 0\textsuperscript{th}-order soft/hard attention models.

On the other hand, our two-step attention trained with the additional copying task already improves upon the baseline without any of the additional biases, with a dev accuracy of $48$ compared to $40.3$ for the best baseline model.
We attribute this to two factors, the first being our novel architecture. Compared to a single attention over the concatenated tag and lemma sequences, our two-step attention has the advantage of two distinct attention mechanisms, which capture the inherently different properties of the tag and the lemma character sequences. Another advantage is the two-step process that guides the lemma attention with the tags. 
Disentangling the two attentions \textit{and} ordering them in an intuitive way makes it easier for them to learn their respective tasks.

The second factor is, we suspect, a slightly better choice of hyperparameters: our models are smaller than the baseline ones. Although we did not tune our hyperparameters extensively, a few experiments on a couple of language pairs showed that increasing the model size hurt performance under these extremely data-scarce conditions.

Each of the additional techniques we tested further contributed a few accuracy points. The attention biases add~$0.7\%$ points and scheduled sampling in the third training phase further adds~$0.4\%$ points.
\ifaclfinal
The different batching schedule with large batch sizes in the beginning but smaller towards the end of training helps a little more in terms of accuracy, but most importantly it significantly speeds up training.
\fi
Table~\ref{tab:model_ablations} also reports the development set accuracy when using hallucinated data. Although the large improvements are also proportionally reflected in the test set, these numbers are not directly comparable to the rest since the development set data were used in the hallucination process.
%In an ideal high-resource scenario one could treat the development set as an additional blind test and not use it for anything but regularization through early stopping, but under such low-resource scenarios we find that taking advantage of every possible additional signal is important.

{
\begin{table}[h!]
\centering
\tiny
%\begin{tabular}{l|r}
\begin{tabular}{ccccccc}
\toprule
L1 & L2 & L1+L2 & $+\mathcal{L}_l$ & $+\mathcal{H}$ & $+\mathcal{L}_l+\mathcal{H}$ & $\mathcal{H}$ \\ \midrule
turkish & \multirow{5}{*}{azeri} & 81 &  77 &  80 &  81 &  \multirow{5}{*}{66.7$\pm$0.9}  \\ 
persian & &  55 &  63 &  74 &  69 &  \\ 
bashkir & &  57 &  59 &  66 &  67 &  \\
uzbek & &  47 &  55 &  74 &  70 &  \\ 
all & & 84 & 71 & 83 & \textbf{87}\\ \midrule
urdu & \multirow{5}{*}{bengali} &  42 &  32 &  66 &  \textbf{67} &  \multirow{5}{*}{63.7$\pm$4.0}  \\
sanskrit & &  44 &  38 &  66 &  65 &  \\ 
hindi & &  49 &  52 &  67 &  65 &  \\ 
greek & &  42 &  46 &  65 &  \textbf{67} & \\ 
all & & 49 & 50 & 64  &   62 \\ \midrule
turkish & \multirow{2}{*}[-.6em]{crimean} &  87 &  80 &  85 &  \textbf{89} & \multirow{4}{*}{71.3$\pm$1.1} \\ 
bashkir & &  59 &  60 & 70 &  69 &  \\ 
uzbek & \multirow{2}{*}[.6em]{tatar} &  60 &  60 & 72 &  67 &  \\ 
all & & 82 & 81 & 88 & 80 & \\ \midrule
finnish & \multirow{4}{*}{ingrian} &  38 &  36 &  34 & 40 &  \multirow{4}{*}{34.6$\pm$2.3}  \\ 
hungarian & &  28 &  32 &  32 & 38 &  \\ 
estonian & &  32 &  32 &  32 & 38 & \\ 
all & & 38 & \textbf{44} & 36 & 36 & \\ \midrule
finnish & \multirow{4}{*}{karelian} &  54 &  50 & 58 &  56 & \multirow{4}{*}{52.6$\pm$1.1}  \\ 
hungarian & &  42 &  46 & 58 &  52 & \\ 
estonian & &  42 &  42 &  54 & 58 & \\
all & & 50 & 46 & 54 & \textbf{64} & \\ \midrule
basque & \multirow{5}{*}{kashubian} &  46 &  40 &  70 &  76 &  \multirow{5}{*}{74.0$\pm$2.3}  \\ 
slovak & &  58 &  62 &  74 & 76 & \\ 
czech & &  54 &  64 & 78 &  66 &  \\ 
polish & &  66 &  78 &  78 &  \textbf{80} & \\ 
all & & 70 & 72 & 78 & \textbf{80} & \\ \midrule
bashkir & \multirow{4}{*}{khakas} &  82 &  78 &  80 & 84 & \multirow{4}{*}{72.6$\pm$6.4}  \\ 
uzbek & & 84 &  80 &  72 &  76 & \\ 
turkish & & 90 & 90 &  80 &  82 &  \\ 
all & & 84 & \textbf{92} & 86 & 84 & \\ \midrule
estonian & \multirow{4}{*}{livonian} &  27 &  27 & 35 &  34 &  \multirow{4}{*}{33$\pm$1} \\ 
hungarian & &  28 &  30 & 35 &  33 &  \\ 
finnish &  &  30 &  26 &  34 & 35 &  \\ 
all & & 26 & 25 & \textbf{36} & \textbf{36} \\ \midrule
italian & \multirow{4}{*}{maltese} &  34 &  25 &  \textbf{48} &  42 &  \multirow{4}{*}{45.6$\pm$5.6} \\ 
arabic & &  18 &  26 & 41 &  38 & \\ 
hebrew & &  20 &  21 & 47 &  45 &   \\ 
all & & 29 & 28 & 40 & 46 \\ \midrule
danish & \multirow{1}{*}[-.4em]{middle} &  68 &  58 &  78 &  80 &  \multirow{4}{*}{76.7$\pm$6.0}  \\ 
dutch & \multirow{1}{*}[-.5em]{high} &  70 &  62 &  74 & 82 &  \\ 
german & \multirow{2}{*}{german} &  72 &  68 &  \textbf{86} &  82 &  \\
all & & 70 & 70 & 84 & 78 \\ \midrule
danish & \multirow{2}{*}[-.6em]{north} &  18 &  25 &  44 &  46 &  \multirow{4}{*}{43$\pm$1} \\ 
dutch &  &  28 &  22 &  \textbf{47} &  46 &  \\
english & \multirow{2}{*}[.6em]{frisian}  &  22 &  26 &  \textbf{47} &  42 & \\ 
all & & 23 & 25 & 43 & 46 & \\ \midrule
asturian & \multirow{4}{*}{occitan} &  58 &  49 &  77 &  74 & \multirow{4}{*}{78.3$\pm$3.5} \\ 
spanish & &  47 &  55 &  78 &  76 &  \\ 
french & &  41 &  52 & \textbf{80}& \textbf{80} &  \\ 
all & & 52 & 61 & 78 & \textbf{80} & \\ \midrule
russian & \multirow{1}{*}[-.4em]{old} &  40 &  39 &  39 &  \textbf{64} &  \multirow{4}{*}{57.3$\pm$1.5} \\ 
polish & \multirow{1}{*}[-.5em]{church} &  41 &  41 & 59 &  58 &  \\ 
bulgarian & \multirow{2}{*}{slavonic} &  38 &  42 &  44 &  56  \\
all & & 41 & 42 & \textbf{64} & 56 & \\ \midrule
sanskrit & \multirow{3}{*}{pashto} &  20 &  16 &  46 &  44 &  \multirow{3}{*}{46.5$\pm$3.5}  \\ 
persian & &  36 &  28 &  39 & 48 &  \\ 
all & & 29 & 30 & 46 & \textbf{49} & \\ \midrule
welsh & \multirow{2}{*}[-.6em]{scottish} &  46 &  34 &  64 & 64 &  \multirow{4}{*}{58$\pm$2}  \\ 
irish & &  58 &  62 &  \textbf{68} &  66 &   \\ 
latvian & \multirow{2}{*}[.6em]{gaelic}  &  58 &  36 &  58 & 66 &   \\ 
all & & 60 & 64 & 62 & 66 & \\ \midrule
bashkir & \multirow{4}{*}{tatar} &  67 &  64 &  73 &  69 &  \multirow{4}{*}{73.0$\pm$4.6} \\ 
uzbek &  &  55 &  45 &  67 & 72 & \\ 
turkish &  &  81 &  79 &  82 &  75  \\ 
all & & 81 & 79 & \textbf{84} & 83 & \\ \midrule
\multicolumn{2}{c}{Average (over all pairs)} & 49.00 & 48.48 & 59.36 & 60.18 & 57.9\\
\bottomrule
\end{tabular}
\caption{Results with multiple transfer languages (sample). The best performing system per target (L2) language is \textbf{highlighted}. We repeated the hallucinated-data-only experiments as many times as potential transfer (L1) languages, hence the $\mathcal{H}$ column reports average accuracy $\pm$ standard deviation.}
\label{tab:results_multiple_one}
\end{table}
}

\section{Analysis}
\label{sec:analysis}

We analyze the results over various groupings of languages to elucidate the properties of our models over the~100 quite diverse language pairs. We use typological information from the URIEL database \cite{littel-et-al:2017} in this analysis.

\paragraph{Single Language Transfer} We first focus on the test languages for which a \textit{single} transfer language was provided, with detailed results presented in
Table~\ref{tab:single_results}. 
Generally, the average genetic distance between the transfer and the task language is quite high $(0.75)$ for these pairs. 
It is easy to observe that the larger the typological distance, the larger the improvement from adding hallucinated data.
In fact, excluding the languages with no typological information available, there is strong correlation~($\rho$=0.6) between genetic distance and improvement from hallucinated data.

For cross-lingual transfer to be useful, the languages need to be at least somewhat related and share similar characteristics. A prime example is transfer from Italian for Neapolitan, which achieves a~70\% accuracy without any additional synthesized data.
In the same vein, the same condition is necessary for the adversarial language discriminator to have impact, as using it on extremely distant language pairs leads to worse performance (e.g. Russian-Portuguese, Bengali-Greek, or Urdu-Old English).
This is expected, as forcing language invariant representations across vastly different languages is analogous to representing a bimodal distribution with its mean.

The results on Kurmanji-Sorani (Northern Kurdish-Central Kurdish) seem to be a valid counter-example to the above statement, i.e. the two languages are related,\footnote{A reviewer pointed out that Kurmanji and Sorani might actually be more distant than the politically-motivated term ``Kurdish dialects" might imply.} but cross-lingual transfer without hallucinated data performs poorly, achieving a mere~16.2 accuracy.
The reason for this discrepancy lies in the characters: Kurmanji is written in the Latin alphabet, while Sorani uses the Arabic one.\footnote{Kurmanji is also sometimes written with the Cyrillic or the Nashk Arabic scripts, but our data are Latin-only.}
The lack of any similar representations across the languages is too hard to overcome even with the adversarial language discriminator.\footnote{We did not attempt to address this issue, by e.g. increasing the loss weight, but we leave this for future work.}

\paragraph{Multiple Transfer Languages} For most low-resource languages and especially dialects, there exist several possible candidate transfer languages that can be related enough to satisfy the similarity constraint.
We present extensive ablations on such cases in Table~\ref{tab:results_multiple_one} with results on the rest of the SIGMORPHON language pairs.\footnote{We present a sample due to space constraints. The discussion pertains to all language pairs. The full table is available in the Appendix.}
%and~\ref{tab:results_multiple_two}, which list the rest of the SIGMORPHON language pairs.

%\input{new-results2.tex}

We again observe positive correlations between the language genetic proximity and the performance of cross-lingual transfer, even with all the transfer and test languages being related. For example, transfer from (distant) Basque to Kashubian performs about~10 percentage points worse than transfer from (related) Slovak or Czech, which in turn perform worse than transfer from Polish (more closely related to Kashubian than the others). Transfer for North Frisian using Danish is also~10 percentage points worse than transfer from the more closely related Dutch.
It is worth noting that using the hallucinated data reduces the effect of the genetic distance between languages, with the standard deviation across the results within the same test language becoming much smaller.

A very interesting case of cross-lingual transfer is Maltese, which is a semitic language and hence genetically close to Hebrew and Arabic. Surprisingly, we obtain better results when transferring from Italian. Again, a script discrepancy could be the main reason, also considering that the root and pattern morphology is only partially expressed in the scripts of Hebrew and Arabic, whereas it is fully expressed (by writing all the vowels) in Maltese.
We should also point out that genetic similarity might not be enlightening enough. As \citet{hoberman2003verbal} point out, the productive verbal morphology in Maltese has become affixal due to borrowings from Romance languages. To further exacerbate the situation, all provided train, dev, and test examples are verbs (no nouns or adjectives), providing an explanation to this seemingly counter-intuitive result. 

Another interesting case is that of cross-lingual transfer for Bengali, with the potential languages varying from very related (Sanskrit, Hindi) to even only distantly related (Greek). Nevertheless, there is notably little variance in the performance of the systems. We believe that again the culprit is the difference in writing systems between all selected transfer and test language, which does not allow our system to leverage cross-lingual information: our Bengali data use the Bengali script, the Urdu dataset is in the Arabic one, Hindi and Sanskrit use Devanagari, and Greek uses the Greek alphabet.

We also present analytic results with cross-lingual transfer from all transfer languages from the suggested SIGMORPHON pairs.\footnote{We do not use \textit{all} languages for transfer in a test language. For instance, when testing on Occitan `all' stands for training on Asturian, Spanish, and French.}
In~14 out of the~29 test languages, our best-performing model is trained on multiple transfer languages. For instance, using Turkish, Persian, Bashkir, and Uzbek data for transfer to Azeri leads to a~6 point improvement over any single-language-transfer result. A potential explanation is that a dialect/language has indeed been influenced by multiple languages. Another reason could lie in the increased amount of data and potential regularization effects. We suspect the truth lies in the union of those factors, but nonetheless we conclude that \textbf{whenever available, transfer from multiple related languages can further improve accuracy}.

In our experiments we used all transfer languages that the SIGMORPHON organizers proposed in the~2019 challenge.
%, but our results could potentially further improve.
On the one hand, a more sophisticated data selection process could likely yield improvements. For example, Yiddish is primarily based in High German and it has elements from Hebrew, Aramaic, as well as Slavic languages, but we did not test how transfer from these languages might perform. Moreover, in order to remain faithful to the SIGMORPHON challenge, we used some distant languages that probably worsen the results (e.g. Greek for Bengali).
%, or Latvian for Scottish Gaelic).

Also, alphabet divergence issues still need to be addressed. For instance, we suspect that our accuracy in Yiddish (which uses the Hebrew alphabet with Yiddish orthography) could be greatly improved by finding some type of mapping between its orthography and the Latin script that most of its related languages use. The same could hold for transfer for the central Asian languages (Tatar, Turkmen, Azeri among others) which use a variety of the Latin, Cyrillic, or Arabic scripts.

Lastly, we experimented with a completely monolingual setting, using just the low-resource and hallucinated language data (columns $\mathcal{H}$ in Tables~\ref{tab:single_results} and~\ref{tab:results_multiple_one}). For fairer comparison to cross-lingual transfer, we repeated the hallucination process as many times as candidate transfer languages and we report the mean and standard deviation of the test set accuracy. This baseline is extremely competitive, lagging only a few points behind the $\mathcal{L}+\mathcal{H}$ combination.
Encouragingly, this entails that hallucination is a viable option for entire language families without a single high-resource representative or low-resource isolates.

\paragraph{Interpretability}

\begin{figure}
    \centering
    \small
    \begin{tabular}{@{}cc@{}}
    Kazakh & Modern Greek \\
    \begin{tabular}{@{}c@{}c@{}c@{}}
    &{\small \foreignlanguage{russian}{а \ с \ п \ а \ н}}&{\tiny \texttt{\: N \,ACC PL}} \\
    %&{ \small \foreignlanguage{greek}{π ρ ο σ α ρ τ ώ}} & \tiny{\texttt{V\,3\,PL\,PFV\,SBJV}} \\ 
    \foreignlanguage{russian}{а}&
    \multirow{9}{*}{\includegraphics[scale=.28]{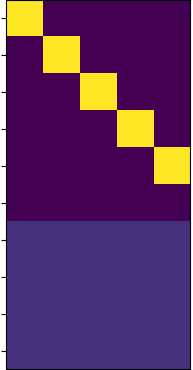}} & 
    \multirow{9}{*}{\includegraphics[scale=.28]{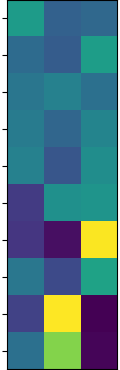}} \\
    \foreignlanguage{russian}{с}&\\[.09em]
    \foreignlanguage{russian}{п}&\\
    \foreignlanguage{russian}{а}&\\[0.1em]
    \foreignlanguage{russian}{н}&\\[-.1em]
    \foreignlanguage{russian}{д}&\\[0.1em]
    \foreignlanguage{russian}{а}&\\[.09em]
    \foreignlanguage{russian}{р}&\\[.09em]
    \foreignlanguage{russian}{д}&\\[.09em]
    \foreignlanguage{russian}{ы}&\\
    &$a_x$ & $a_t$\\[-1em]
    \end{tabular}
    &
    \begin{tabular}{@{}c@{}c@{}c@{}}
    &{\small \foreignlanguage{greek}{π\:ρ\:ο\:σ\:α\:ρ\:τ\:ώ}} & \tiny{\texttt{V\,3\,Pl\,Pfv\,Sbjv}} \\[-.5em]
    &
    \multirow{12}{*}{\includegraphics[scale=.28]{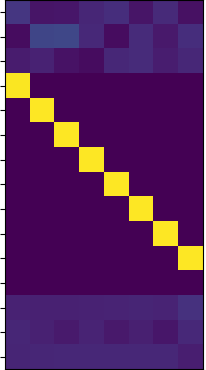}} & 
    \multirow{12}{*}{\includegraphics[scale=.28]{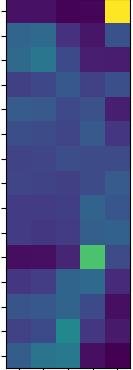}} \\[-.4em]
    \small{\foreignlanguage{greek}{ν}}&\\[-.4em]
    \small{\foreignlanguage{greek}{α}}&\\[.3em]
    \small{\foreignlanguage{greek}{π}}&\\[-.4em]
    \small{\foreignlanguage{greek}{ρ}}&\\[-.4em]
    \small{\foreignlanguage{greek}{ο}}&\\[-.4em]
    \small{\foreignlanguage{greek}{σ}}&\\[-.4em]
    \small{\foreignlanguage{greek}{α}}&\\[-.4em]
    \small{\foreignlanguage{greek}{ρ}}&\\[-.4em]
    \small{\foreignlanguage{greek}{τ}}&\\[-.4em]
    \small{\foreignlanguage{greek}{ή}}&\\[-.4em]
    \small{\foreignlanguage{greek}{σ}}&\\[-.4em]
    \small{\foreignlanguage{greek}{ο}}&\\[-.4em]
    \small{\foreignlanguage{greek}{υ}}&\\[-.4em]
    \small{\foreignlanguage{greek}{μ}}&\\[-.4em]
    \small{\foreignlanguage{greek}{ε}}&\\
    &$a_x$ & $a_t$\\[-1em]
    \end{tabular}
    \end{tabular}
    \caption{Attention visualization examples.
    \ifaclfinal 
    The inflected form is generated from top to bottom.
    \fi}
    \label{fig:twoexample}
\end{figure}

We find that the attention matrices can help understand our model's predictions. A visualization of two examples is shown in Figure~\ref{fig:twoexample}, showcasing the interpretability advantage of the disentangled two-step attentions. In the Kazakh example the tag attention clearly identifies the suffixes \foreignlanguage{russian}{да} (that marks plural) and \texttt{\foreignlanguage{russian}{рды}} (that marks the accusative case). The Greek example is a great one of how the two-step process allows the tags to guide the lemma attention. Due to the
%subjunctive 
{\small\texttt{SBJV}} tag, the model does not use the lemma until the necessary particle \foreignlanguage{greek}{να} has been generated. Consequently, the lemma attention properly copies the stem, and then the tag attention attends first over {\small\texttt{PFV}} and then over {\small\texttt{PL}} and~{\small\texttt{3}} in order to construct the correct 
%series of suffixes 
suffix
for perfective and~3\textsuperscript{rd} person plural.

\iffalse
\begin{enumerate}
    \item Greek: test data sometimes do not include crucial inherent morphology information. For example: (todo: find the actual instance) an ambiguous noun that could be either female or neutral based on its ending. The tags did not include gender information, so the model inflected \texttt{PL;ACC} as if it was female (more common in training data). When providing the full information (by adding the correct  gender tag \texttt{NEUT}) the \gnc{}{model} was able to correctly inflect it.
\end{enumerate}
\fi

\iffalse
\input{figures/greekngram.tex}
Why does data hallucination help? Mitigates exposure bias. See Figure~\ref{fig:greekngram}.

Further Analysis (todo):
\begin{itemize}
    \item Use labels on languages (fusional, agglutinative) --no polysynthetic ones in the dataset :( -- and see if there's any significant difference in performance.
    \item The same, but over pairs.
\end{itemize}
\fi

\section{Related Work}
\label{sec:related}
The inflection task in high-resource settings has been extensively studied through the SIGMORPHON shared tasks. Notably, the best models
explicitly model copying and hard monotonic attention \cite{aharoni2016improving,aharoni2017morphological} with the previous state-of-the-art forcing strict monotonicity
\cite{wu-2019-exact}.
We instead achieve state-of-the-art with a cheaper approach that simply intermixes a copying task which also encourages monotonicity.

Data augmentation for inflection has been explored by \citet{bergmanis2017training} and \citet{zhou2017morphological} among others.
The work of \citet{silfverberg2017data} is the most similar to ours, but as we already discussed, it has a few shortcomings that our approach addresses.

\citet{kann-etal-2017-one} have identified typology as playing a role for cross-lingual transfer, but they measure language similarity using lexical overlap. We attest that this data-based measure is less informative and more suspect to variation, so we instead use the genetic typological information to quantify correlations between performance improvements and language distance.

Our novel two-step process decoder architecture bares similarities with multi-source models
\cite{anastasopoulos+chiang:interspeech2018,zoph2016multisource} which provide two contexts from two encoded sources to the decoder. A similar disentangled encoding was also used by \citet{acs-2018-bme} for their SIGMORPHON 2018 submission. We in fact experimented with this architecture but preliminary results on the development sets showed that our two-step architecture achieved better performance.
Interestingly, the second-best performing system \cite{peters-martins-2019-ist} at SIGMORPHON 2019, which also ranked first in terms of Levenshtein distance, also uses decoupled encoders to separately encode the lemma and the tags; this further cosolidates our belief that such an approach is superior to using a single encoder for the concatentated sequence of the tags and lemma. The main difference to our model is that they do not use our two-step decoder process, while they substitute all softmax operations with sparsemax \cite{martins2016softmax}, yielding interpretable attention matrices very similar to ours.
The use of sparsemax in conjunction with our two-step decoder process, as well as along our data hallucination technique, presents a promising direction towards even better results in the future.

%Contextualizing morphological inflection:
%\cite{vylomova2019contextualization}

\section{Conclusion}

With this work we advance the state-of-the-art for morphological inflection on low-resource languages by~15 points, through a novel architecture, data hallucination, and a variety of training techniques. Our two-step attention decoder follows an intuitive order, also enhancing interpretability.
We also suggest that complicated methods for copying and forcing monotonicity are unnecessary. We identify language genetic similarity as a major success factor for cross-lingual training, and show that using many related languages leads to even better performance.
Despite this significant stride, the problem is far from solved.
Language-specific or language-family-specific improvements (i.e. proper dealing with different alphabets, or using an adversarial language discriminator) could potentially further boost performance.

%Although we only test it in the inflection task, this two-step approach could potentially be useful for any task that requires controlled language generation. Such examples are aspect-based summarization and domain- or aspect-specific translation, where a set of additional input tags describes the desired properties of the output.

\section*{Acknowledgements}
The authors are grateful to the anonymous reviewers for their exceptionally constructive and insightful comments, to Arya McCarthy for discussions on morphological inflection, as well as to Gabriela Weigel for her invaluable help with editing and proofreading the paper. 
This material is based upon work generously supported by the National Science Foundation under grant 1761548.

\bibliographystyle{acl_natbib}
\bibliography{References}

\appendix
\section{Hyperparameters}
Here we list all the hyperparameters of our models:
\begin{itemize}[noitemsep]
    \item Encoder/Decoder number of layers: 1
    \item Character/Tag Embedding size: 32
    \item Recurrent State Size: 100
    \item LSTM Type: CoupledLSTM
    \item Attention Size: 100
    \item Target and Source Character Embeddings are Tied
    \item Tag Self-Attention size: 100
    \item Tag Self-Attention heads: 1
\end{itemize}
Here we list the optimizer settings:
\begin{itemize}[noitemsep]
    \item Optimizer: SimpleSGD
    \item Starting learning rate: 0.1
    \item Learning Rate Decay: 0.5
    \item Learning Rate Decay Patience: 6 epochs
    \item Maximum number of epochs: 20, 40, 40 (for each training phase)
    \item Minibatch Size: 10, 10, 2 (for each training phase).
\end{itemize}

\onecolumn
\pagebreak

\section{Complete Result Tables}

Table~\ref{tab:single_results_app} lists all results for test languages with a single candidate transfer language. Results on test languages with multiple candidate languages (both single-language-transfer, and transfer with all candidate suggested languages) are listed in Tables~\ref{tab:results_multiple_one_app} and~\ref{tab:results_multiple_two_app}.

\begin{table*}[t]
\centering
\small
\begin{tabular}{cccccccc}
\toprule
L1 & L2 & Genetic dist. & L1+L2 & $+\mathcal{L}_l$ & $+\mathcal{H}$ & $+\mathcal{L}_l+\mathcal{H}$ & $\mathcal{H}$ \\ \midrule
latin & czech & 0.86 &  15.0 &  26.0 &  71.4 &  68.0 &  \textbf{77.4} \\ 
bengali & greek & 0.88 &  22.4 &  16.4 &  70.5 &  70.6 &  \textbf{71.6} \\
sorani & irish & 1.0 &  20.3 & 18.6 &  \textbf{66.3} &  64.6 &  65.6 \\
italian & ladin & 0.30 &  48 &  54 &  \textbf{74} &  \textbf{74} &  \textbf{74}  \\
latvian & lithuanian & 0.25 &  17.1 &  23.2 &  48.4 &  48.4 &  \textbf{50.5}  \\
english & murrinhpatha & N/A &  \textbf{36} &  2 &  6 &  20 &  20  \\
italian & neapolitan & 0.10 & 70 &  70 &  83 &  83 &  \textbf{84} \\ 
urdu & old english & 0.88 & 23.8 &  18.5 &  43.4 &  40.4 &  \textbf{44.3} \\
slovene & old saxon & 0.89 & 10.7 &  14.7 &  \textbf{52.3} &  50.5 &  50.5  \\
russian & portuguese & 0.93 & 34.5 &  22.2 &  \textbf{88.8} &  88.4 &  87.7  \\
swahili & quechua & 1.0 & 14.2 &  12.8 &  \textbf{92.1} &  91.6 &  91.6  \\ 
portuguese & russian & 0.93 &  25.6 &  17.5 &  \textbf{76.3} &  74.6 &  74.3  \\
kurmanji & sorani & N/A & 16.2 &  13.6 &  \textbf{69.0} &  64.3 &  66.7 \\
zulu & swahili & 1.0 & 46 &  52 &  \textbf{81} &  80 &  76 \\
kannada & telugu & 0.75 & 76 &  72 &  \textbf{94} &  90 &  \textbf{94} \\ \midrule
\multicolumn{2}{c}{Average} & 0.75 & 31.72 & 28.90 & 67.77 & 67.23 & 68.55 \\
\bottomrule
\end{tabular}
\caption{Results with a single transfer language. Monolingual data hallucination is crucial due to the distance of the languages. In some cases cross lingual transfer should be avoided in favor of a purely monolingual setting ($\mathcal{H})$.}
\label{tab:single_results_app}
\end{table*}

\twocolumn

\begin{table}[t!]
\centering
\tiny
%\begin{tabular}{l|r}
\begin{tabular}{ccccccc}
\toprule
L1 & L2 & L1+L2 & $+\mathcal{L}_l$ & $+\mathcal{H}$ & $+\mathcal{L}_l+\mathcal{H}$ & $\mathcal{H}$ \\ \midrule
turkish & \multirow{5}{*}{azeri} & 81 &  77 &  80 &  81 &  \multirow{5}{*}{66.7$\pm$0.9}  \\ 
persian & &  55 &  63 &  74 &  69 &  \\ 
bashkir & &  57 &  59 &  66 &  67 &  \\
uzbek & &  47 &  55 &  74 &  70 &  \\ 
all & & 84 & 71 & 83 & \textbf{87}\\ \midrule
urdu & \multirow{5}{*}{bengali} &  42 &  32 &  66 &  \textbf{67} &  \multirow{5}{*}{63.7$\pm$4.0}  \\
sanskrit & &  44 &  38 &  66 &  65 &  \\ 
hindi & &  49 &  52 &  67 &  65 &  \\ 
greek & &  42 &  46 &  65 &  \textbf{67} & \\ 
all & & 49 & 50 & 64  &   62 \\ \midrule
welsh & \multirow{4}{*}{breton} &  54 &  55 &  83 &  86 &  \multirow{4}{*}{78.2$\pm$2.5}  \\ 
irish & &  48 &  52 &  80 &  \textbf{88} &  \\
albanian & &  53 &  59 &  78 &  81 &  \\
all & & 62 & 55 & 81 & 82 & \\ \midrule
hebrew & \multirow{1}{*}[-0.7em]{classical} &  86 &  85 & 95 & 95 & \multirow{3}{*}{94$\pm$0} \\ 
arabic & \multirow{2}{*}{syriac} &  83 &  82 &  92 & 95 &  \\ 
all & & 87 & 85 & 95 & \textbf{96} & \\ \midrule
irish & \multirow{3}{*}{cornish} &  18 &  24 & 24 &  20 & \multirow{3}{*}{24$\pm$0} \\ 
welsh & &  26 &  \textbf{28} &  \textbf{28} &  24 & \\
all & & 22 & 16 & 22 & 24 & \\ \midrule
turkish & \multirow{2}{*}[-.6em]{crimean} &  87 &  80 &  85 &  \textbf{89} & \multirow{4}{*}{71.3$\pm$1.1} \\ 
bashkir & &  59 &  60 & 70 &  69 &  \\ 
uzbek & \multirow{2}{*}[.6em]{tatar} &  60 &  60 & 72 &  67 &  \\ 
all & & 82 & 81 & 88 & 80 & \\ \midrule
spanish & \multirow{3}{*}{friulian} &  55 &  55 &  81 &  \textbf{83} & \multirow{3}{*}{79.0$\pm$2.6} \\ 
italian & &  56 &  55 &  78 &  77 &  \\ 
all & & 64 & 58 & 78 & \textbf{83} & \\ \midrule
finnish & \multirow{4}{*}{ingrian} &  38 &  36 &  34 & 40 &  \multirow{4}{*}{34.6$\pm$2.3}  \\ 
hungarian & &  28 &  32 &  32 & 38 &  \\ 
estonian & &  32 &  32 &  32 & 38 & \\ 
all & & 38 & \textbf{44} & 36 & 36 & \\ \midrule
adyghe & \multirow{3}{*}{kabardian} &  92 &  92 &  93 &  \textbf{96} &  \multirow{3}{*}{90.6$\pm$4.0} \\ 
armenian & &  78 &  77 &  86 &  80 & \\ 
all & & 95 & 91 & 93 & 90 & \\ \midrule
finnish & \multirow{4}{*}{karelian} &  54 &  50 & 58 &  56 & \multirow{4}{*}{52.6$\pm$1.1}  \\ 
hungarian & &  42 &  46 & 58 &  52 & \\ 
estonian & &  42 &  42 &  54 & 58 & \\
all & & 50 & 46 & 54 & \textbf{64} & \\ \midrule
basque & \multirow{5}{*}{kashubian} &  46 &  40 &  70 &  76 &  \multirow{5}{*}{74.0$\pm$2.3}  \\ 
slovak & &  58 &  62 &  74 & 76 & \\ 
czech & &  54 &  64 & 78 &  66 &  \\ 
polish & &  66 &  78 &  78 &  \textbf{80} & \\ 
all & & 70 & 72 & 78 & \textbf{80} & \\ \midrule
turkish & \multirow{4}{*}{kazakh} &  \textbf{86} &  80 &  74 &  66 &  \multirow{4}{*}{73.3$\pm$3.0} \\ 
bashkir &  &  68 & 84 &  74 &  70 &  \\ 
uzbek & &  66 &  60 & 78 &  70 &   \\ 
all & & 76 & 78 & 80 & 78 & \\ \midrule
bashkir & \multirow{4}{*}{khakas} &  82 &  78 &  80 & 84 & \multirow{4}{*}{72.6$\pm$6.4}  \\ 
uzbek & & 84 &  80 &  72 &  76 & \\ 
turkish & & 90 & 90 &  80 &  82 &  \\ 
all & & 84 & \textbf{92} & 86 & 84 & \\ \midrule
czech & \multirow{3}{*}{latin} &  5.4 &  5.0 &  20.6 & 42.0 &  \multirow{3}{*}{41.9$\pm$1.4}  \\ 
romanian & &  7.9 &  9.0 &  18.8 &  41.3 &\\ 
all & & 3.6 &  7.7 & 40.1 &  \textbf{44.1} \\ \midrule
estonian & \multirow{4}{*}{livonian} &  27 &  27 & 35 &  34 &  \multirow{4}{*}{33$\pm$1} \\ 
hungarian & &  28 &  30 & 35 &  33 &  \\ 
finnish &  &  30 &  26 &  34 & 35 &  \\ 
all & & 26 & 25 & \textbf{36} & \textbf{36} \\ 
\bottomrule
\end{tabular}
\caption{Multiple transfer language results (part 1).}
\label{tab:results_multiple_one_app}
\end{table}
\begin{table}[t]
\centering
\tiny
\begin{tabular}{cccccccc}
\toprule
L1 & L2 & L1+L2 & $+\mathcal{L}_l$ & $+\mathcal{H}$ & $+\mathcal{L}_l+\mathcal{H}$ & $\mathcal{H}$ \\ \midrule
italian & \multirow{4}{*}{maltese} &  34 &  25 &  \textbf{48} &  42 &  \multirow{4}{*}{45.6$\pm$5.6} \\ 
arabic & &  18 &  26 & 41 &  38 & \\ 
hebrew & &  20 &  21 & 47 &  45 &   \\ 
all & & 29 & 28 & 40 & 46 \\ \midrule
danish & \multirow{1}{*}[-.4em]{middle} &  68 &  58 &  78 &  80 &  \multirow{4}{*}{76.7$\pm$6.0}  \\ 
dutch & \multirow{1}{*}[-.5em]{high} &  70 &  62 &  74 & 82 &  \\ 
german & \multirow{2}{*}{german} &  72 &  68 &  \textbf{86} &  82 &  \\
all & & 70 & 70 & 84 & 78 \\ \midrule
dutch & \multirow{1}{*}[-.4em]{middle} &  36 & 42 &  32 &  36 &  \multirow{4}{*}{36.6$\pm$3.0} \\
german & \multirow{1}{*}[-.6em]{low} &  \textbf{46} &  26 &  32 &  38 & \\ 
danish & \multirow{2}{*}{german} &  30 &  24 &  30 &  30 & \\
all & & 32 & 42 & 36 & 38 & \\ \midrule
danish & \multirow{2}{*}[-.6em]{north} &  18 &  25 &  44 &  46 &  \multirow{4}{*}{43$\pm$1} \\ 
dutch &  &  28 &  22 &  \textbf{47} &  46 &  \\
english & \multirow{2}{*}[.6em]{frisian}  &  22 &  26 &  \textbf{47} &  42 & \\ 
all & & 23 & 25 & 43 & 46 & \\ \midrule
asturian & \multirow{4}{*}{occitan} &  58 &  49 &  77 &  74 & \multirow{4}{*}{78.3$\pm$3.5} \\ 
spanish & &  47 &  55 &  78 &  76 &  \\ 
french & &  41 &  52 & \textbf{80}& \textbf{80} &  \\ 
all & & 52 & 61 & 78 & \textbf{80} & \\ \midrule
russian & \multirow{1}{*}[-.4em]{old} &  40 &  39 &  39 &  \textbf{64} &  \multirow{4}{*}{57.3$\pm$1.5} \\ 
polish & \multirow{1}{*}[-.5em]{church} &  41 &  41 & 59 &  58 &  \\ 
bulgarian & \multirow{2}{*}{slavonic} &  38 &  42 &  44 &  56  \\
all & & 41 & 42 & \textbf{64} & 56 & \\ \midrule
irish & \multirow{4}{*}{old irish} &  2 &  4 &  \textbf{12} &  2 &  \multirow{4}{*}{6$\pm$2} \\ 
belarusian & &  10 & 10 & 10 &  6 &  \\
welsh &  &  4 &  6 & 10 &  6 &   \\ 
all & & 4 & 4 & 8 & 6 & \\ \midrule
sanskrit & \multirow{3}{*}{pashto} &  20 &  16 &  46 &  44 &  \multirow{3}{*}{46.5$\pm$3.5}  \\ 
persian & &  36 &  28 &  39 & 48 &  \\ 
all & & 29 & 30 & 46 & \textbf{49} & \\ \midrule
welsh & \multirow{2}{*}[-.6em]{scottish} &  46 &  34 &  64 & 64 &  \multirow{4}{*}{58$\pm$2}  \\ 
irish & &  58 &  62 &  \textbf{68} &  66 &   \\ 
latvian & \multirow{2}{*}[.6em]{gaelic}  &  58 &  36 &  58 & 66 &   \\ 
all & & 60 & 64 & 62 & 66 & \\ \midrule
bashkir & \multirow{4}{*}{tatar} &  67 &  64 &  73 &  69 &  \multirow{4}{*}{73.0$\pm$4.6} \\ 
uzbek &  &  55 &  45 &  67 & 72 & \\ 
turkish &  &  81 &  79 &  82 &  75  \\ 
all & & 81 & 79 & \textbf{84} & 83 & \\ \midrule
bashkir & \multirow{5}{*}{turkmen} &  82 &  76 &  82 &  88 & \multirow{5}{*}{80$\pm$4} \\ 
arabic &  &  64 &  66 & 84 &  80 \\ 
turkish & &  82 &  88 &  90 &  \textbf{92}\\ 
uzbek & &  66 &  74 & 86 &  78 &\\ 
all & & \textbf{92} & 90 & 84 & 88 & \\ \midrule
estonian & \multirow{4}{*}{votic} &  12 &  17 &  23 &  27 &\multirow{4}{*}{26.3$\pm$2.1} \\ 
finnish &  &  24 &  20 &  26 & 28 \\ 
hungarian & &  10 &  10 &  24 &  \textbf{30} \\
all & & 17 & 20 & 28 & 29 & \\ \midrule
dutch & \multirow{2}{*}[-.6em]{west} &  36 &  39 &  44 &  \textbf{49} & \multirow{4}{*}{47$\pm$1}  \\ 
english &  &  36 &  25 &  45 &  43 \\
danish & \multirow{2}{*}[.6em]{frisian} &  33 &  33 &  42 &  43  \\ 
all & & 40 & 38 & 43 & 45 & \\ \midrule
danish & \multirow{4}{*}{yiddish} &  50 &  54 &  55 & 56 & \multirow{4}{*}{55.3$\pm$0.6}  \\
dutch &  &  49 &  50 & 56 &  55 & \\
german & &  53 &  54 &  \textbf{57} &  54  \\
all & & 55 & 55 & 55 & 54 & \\ 
\midrule
\multicolumn{2}{c}{Average (over all pairs)} & 49.00 & 48.48 & 59.36 & 60.18 & 57.9\\
\bottomrule
\end{tabular}
\caption{Multiple transfer language results (part 2).}
\label{tab:results_multiple_two_app}
\end{table}

\end{document}